\documentclass{article}

\usepackage[preprint,nonatbib]{neurips_2025}

\usepackage{orcidlink}

\usepackage[utf8]{inputenc} 
\usepackage[T1]{fontenc}    
\usepackage{url}            
\usepackage{booktabs}       
\usepackage{amsfonts}       
\usepackage{nicefrac}       
\usepackage{microtype}      
\usepackage{xcolor}         
\usepackage{multirow}
\usepackage{graphicx}
\usepackage{soul}
\usepackage{amsmath}
\usepackage{amssymb}
\usepackage{enumitem}
\usepackage[font=small]{caption}
\usepackage[ruled,vlined]{algorithm2e}
\usepackage{array}
\newcolumntype{R}[1]{>{\raggedleft\let\newline\\\arraybackslash\hspace{0pt}}m{#1}}

\SetAlCapFnt{\footnotesize}
\SetAlCapNameFnt{\footnotesize}
\SetAlCapNameSty{}
\usepackage{amssymb}
\usepackage{dsfont}
\usepackage{pifont}
\newcommand{\xmark}{\ding{55}}
\SetKwComment{Comment}{/* }{ */}
\usepackage[table]{xcolor}  
\usepackage{siunitx}  
\definecolor{myred}{RGB}{253,179,179}
\definecolor{myyellow}{RGB}{255,255,185}
\usepackage{tcolorbox}
\newcommand{\myrightcomment}[1]{\hfill\smash{\raisebox{28.0ex}{$\triangleright$~#1}}}
\newcommand{\myrightcommentt}[1]{\hfill\smash{\raisebox{14ex}{$\triangleright$~#1}}}
\usepackage{makecell}
\definecolor{MyLightGreen}{RGB}{235, 245, 235}

\title{Instance Data Condensation for Image Super-Resolution}

\author{%
  Tianhao Peng$\dagger$, Ho Man Kwan$\dagger$, Yuxuan Jiang$\dagger$, Ge Gao$\dagger$, Fan Zhang$\dagger$\\ \textbf{Xiaozhong Xu}$\ddagger$, \textbf{Shan Liu}$\ddagger$, \textbf{David Bull}$\dagger$ \\
  $\dagger$ Visual Information Lab, University of Bristol, UK \\
  $\ddagger$ Tencent Media Lab, Palo Alto, USA.\\
  \texttt{\{tianhao.peng, hm.kwan, yuxuan.jiang, ge1.gao, fan.zhang, dave.bull\}@bristol.ac.uk}, \\ \texttt{\{xiaozhongxu, shanl\}@tencent.com}
}

\begin{document}

\maketitle

\begin{abstract}

Deep learning based Image Super-Resolution (ISR) relies on large training datasets to optimize model generalization; this requires substantial computational and storage resources during training. While dataset condensation (DC) has shown potential in improving data efficiency for high-level computer vision tasks, adopting these methods for ISR is not straightforward due to the different requirements of ISR training, including the use of unlabeled datasets and high resolution images with fine details. In this paper, we propose a novel \textbf{I}nstance \textbf{D}ata \textbf{C}ondensation \textbf{(IDC)} framework specifically for ISR, which achieves data condensation in a per-image manner, aiming to address the limitations when directly applying existing DC methods to the ISR task. Furthermore, the IDC framework is based on a novel Random Local Fourier Feature Extraction and Multi-level Feature Distribution Matching methods, which are designed to generate high-quality synthesized content by aligning its feature distributions with those of the original high-resolution training samples at both global and local levels. This framework has been utilized to condense the most commonly used training dataset for ISR, DIV2K, with a \textbf{10\%} condensation rate. The resulting synthetic dataset offers comparable performance to the original full dataset and excellent training stability when used to train various popular ISR models. To the best of our knowledge, this is the first time that a condensed/synthetic dataset (with a 10\% data volume) has demonstrated such performance. 
\end{abstract}

\section{Introduction}

Image super-resolution (ISR) is a well-established research area in low-level computer vision, which aims to up-sample a low-resolution image to higher resolutions, while recovering fine spatial details. In recent years, deep learning inspired methods \cite{liang2021swinir, jiang2024hiif} have been dominant in this field, offering significant improvements over conventional ISR methods based on classic signal processing theory. These learning-based solutions are typically optimized offline with a large training dataset and deployed online for processing arbitrary input images. The training data is thus crucial for maintaining the model generalization ability and for avoiding overfitting issues. 

To this end, it is common practice to simply increase the amount of training material, but this introduces two primary issues. (i) Training efficiency: a large amount of training content inevitably leads to higher training costs with longer training time and greater storage/memory requirements \cite{paul2021deep}. Although efforts have been made to accelerate the training process by using large batch sizes, these cannot fully reduce training costs due to increased memory consumption \cite{lin2022revisiting}. (ii) Data quality: increased data volume does not guarantee performance improvement - large datasets can be associated with unbalanced content distributions (or bias) and data redundancy, which may result in suboptimal inference performance on certain content and reduced model generalization \cite{zhao2021dataset}. Moreover, privacy concerns can also arise when using large-scale data, as models can inadvertently memorize sensitive information, making them vulnerable to membership inference attacks that could potentially expose training data details \cite{melis2019exploiting,lyu2020threats}.

To address these problems, various dataset refinement approaches have been proposed, including coreset selection \cite{phillips2017coresets,katharopoulos2018not} and dataset pruning \cite{ding2023not, moser2024study}, which result in a smaller, representative subset, derived from the large training dataset based on gradient or deep feature statistics. However, these content selection/pruning methods are constrained by the characteristics of the original content, and hence they cannot achieve optimal performance compared to the large dataset, in particular when the subset is much smaller than the original. It is also noted that dataset distillation \cite{wang2018dataset} and condensation \cite{zhao2021dataset, wang2025NCFM} techniques have been proposed recently for high-level computer vision tasks (e.g., image classification with ground truth labels). These are designed to distill/condense a large dataset into a small but ``informative'' version with \textit{synthetic} content. The aim is to achieve improved training efficiency, comparable model generalization ability (with the original dataset), and enhanced data privacy \cite{chen2022private, dong2022privacy}. 
However, these techniques cannot be directly applied to low-level vision tasks, such as ISR. This is due to the fact that existing methods typically distill images by classes, while low-level vision training samples, such as low/high resolution image pairs for ISR, are typically unlabeled. Furthermore, high-level computer vision tasks prioritize global semantic information rather than capturing the fine-grained, high-frequency textures that are essential for super-resolution. Although there have been early attempts that focus on ISR dataset condensation \cite{zhang2024gsdd, dietz2025study}, these exhibit relatively poor performance or can only be deployed in limited scenarios.

In this context, this paper proposes a new \textbf{Instance Data Condensation} (IDC) framework specifically for image super-resolution, which can significantly reduce the amount of training material and speed up the training process, while maintaining model performance. Taking all the cropped patches from each individual image (instance) in a large dataset as input, this framework generates a small number of synthetic low-resolution training patches with \textit{condensed} information based on \textbf{Random Local Fourier Features} and \textbf{Multi-level Feature Distribution Matching}. These methods are designed to retain the distribution of the local features in the original patches and to ensure the fidelity and diversity of the synthesized patches. These synthetic low-resolution patches are then up-sampled by a pre-trained ISR model to obtain their high-resolution counterparts. Designed for ISR, this framework has been utilized to condense the standard training dataset in the ISR, DIV2K \cite{agustsson2017ntire}, and the larger scale Flickr2K dataset\cite{lim2017enhanced}, resulting in synthetic datasets with only 10\% and 1\% training patches, respectively. The condensed dataset was then used to train three popular ISR network architectures, EDSR \cite{lim2017enhanced}, SwinIR \cite{liang2021swinir} and MambaIRv2 \cite{guo2024mambairv2}, which achieve comparable performance to the same networks trained with the full dataset. Importantly, our condensed data enables significantly faster training convergence (up to 4$\times$). In addition, our method can also be generalized to other low-level vision tasks such as image denoising. The main contributions of this work are summarized below.

\begin{figure}[t]
    \centering
    \begin{minipage}[t]{0.66\linewidth}
        \centering
        \includegraphics[width=\linewidth]{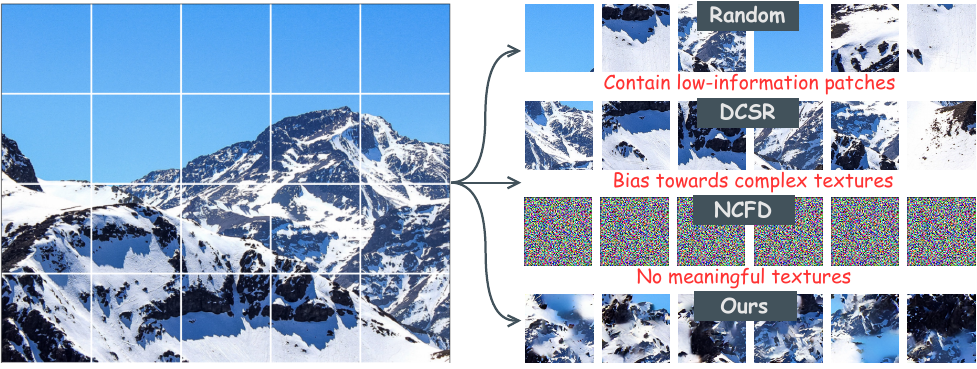}
    \end{minipage}
    \hfill
    \begin{minipage}[t]{0.32\linewidth}
        \centering
        \includegraphics[width=\linewidth]{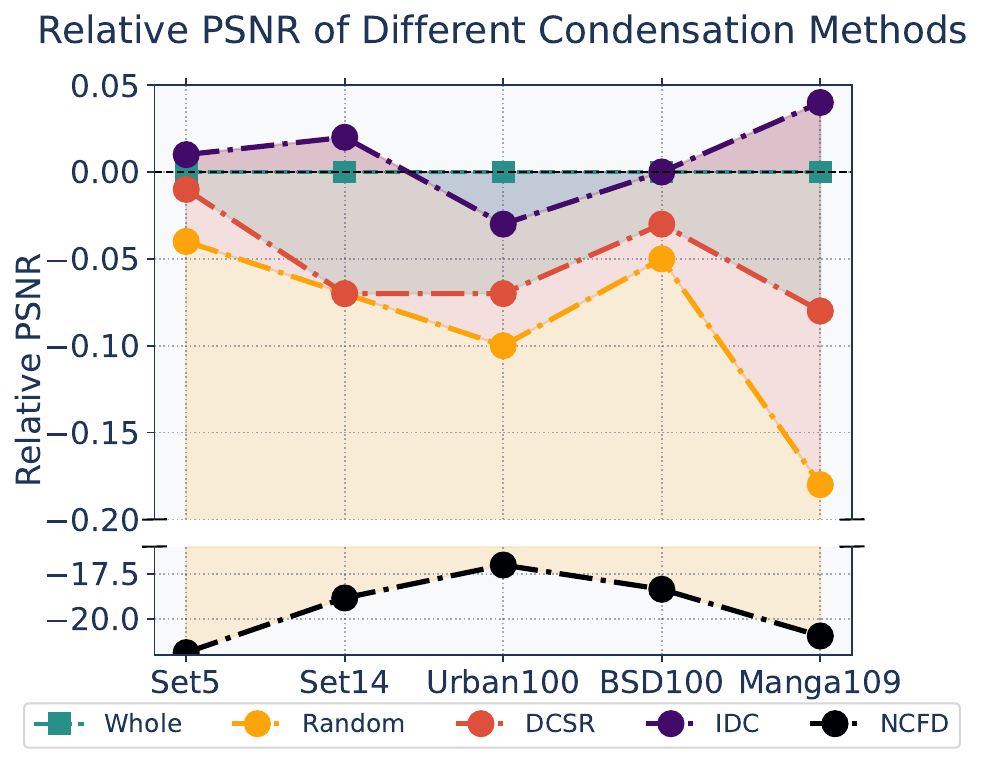}
    \end{minipage}
    \caption{\textbf{(Left)} Visual comparison between synthetic patches generated by our IDC framework and those selected/synthesized by Random Selection, DCSR \cite{ding2023not}, and NCFD (v1 in our ablation study) \cite{wang2025NCFM}. \textbf{(Right)} It shows that an IDC-synthesized dataset (10\% volume) can outperform the full DIV2K dataset when training ISR models.}
    \label{fig:IDC_feature}
    \vspace{-13pt}
\end{figure}

\begin{enumerate}[leftmargin=*]
    \item We propose a new data condensation framework specifically for ISR that operates at the \textbf{instance (image) level}, a paradigm that effectively bypasses the need for class labels common in high-level vision tasks. 
    
    \item We design a \textbf{Multi-level Feature Distribution Matching} approach, which learns feature distributions at the instance and group levels. This hierarchical strategy progressively refines the synthetic data, enhancing feature quality and diversity and leading to well-conditioned samples, enabling high-quality visual feature learning in distribution matching, as shown in \autoref{fig:IDC_feature}.(\textbf{Left}).

    \item We develop the novel \textbf{Random Local Fourier Features}, which capture high-frequency details and local features to facilitate distribution matching in learning high-fidelity synthetic images.

\end{enumerate}

Extensive experimental results demonstrate that our method not only consistently outperforms existing database selection/condensation approaches when used to condense the most commonly used training dataset, DIV2K, for image super resolution, but also significantly \textbf{accelerates training convergence}. By using this approach, a highly condensed dataset can even achieve similar performance to the original full dataset when used for training ISR models, as shown in \autoref{fig:IDC_feature}.(\textbf{Right}). We have further validated its \textbf{generalization} on larger datasets and other low-level tasks (e.g., denoising).

\section{Related Work}
\textbf{Image Super-Resolution (ISR)} is a fundamental task in low-level computer vision that aims to reconstruct high-resolution (HR) images from their low-resolution (LR) counterparts. Over the past decade, advances in deep learning have significantly improved ISR performance, employing ISR models based on network architectures including convolutional neural networks (CNNs) \cite{kim2016accurate, zhang2018image}, vision transformers \cite{wang2022uformer,jiang2025c2d}, and structured state-space models \cite{guo2024mambair,shi2025vmambair}. These ISR methods, due to their data-driven nature, are typically trained on a large training dataset and then deployed for real-world applications. In this case, the training dataset is critical to model performance and generalization. To facilitate ISR model training, a series of datasets have been developed, including small ones such as T91 \cite{yang2010image} and BSD200 \cite{martin2001database}, which contain 91 and 200 natural images, respectively. A significant milestone was the release of DIV2K \cite{agustsson2017ntire}, consisting of 800 high-resolution images, which is now the most commonly used training dataset for ISR. Other alternatives also exist, including Flickr2K \cite{timofte2017ntire}, which comprises 2,650 high-resolution images collected from online sources, and LSDIR \cite{li2023lsdir} with more than 85,000 images. In this paper, we employ DIV2K as the original training dataset to demonstrate the data condensation process due to its popularity.

\noindent\textbf{Dataset condensation and distillation} aim to condense or distill a large-scale original training set into a smaller synthetic one, capable of offering comparable performance on unseen test data when used to train a model for a downstream task. Based on the optimization objectives, current dataset condensation methods can be categorized into three classes: performance matching, parameter matching, and distribution matching. \textit{Performance matching} based approaches \cite{wang2018dataset, zhou2022dataset} are designed to optimize the synthetic dataset to achieve minimum training loss (for the downstream task) compared to using the original training dataset, while \textit{parameter matching} \cite{yu2024teddy} encourages models trained on the synthetic dataset to maintain consistency in the parameter space with those trained on the original dataset. Both types of approaches are similar to bi-level meta-learning \cite{finn2017model} and gradient-based hyperparameter optimization techniques \cite{du2023minimizing},  which are associated with high computation and memory costs and are difficult to apply to high resolution and large scale datasets \cite{cazenavette2022dataset,yu2023dataset}. \textit{Distribution matching} generates synthetic data by directly optimizing the distribution distance between synthetic and real data, which typically leverages feature embeddings extracted from networks with different initializations to construct the distribution space, utilizing MMD as the distribution metric \cite{zhao2023improved}. This has been further enhanced by incorporating batch normalization statistics \cite{yin2023squeeze, du2024diversity, shao2024generalized} or neural features in the complex plane
 \cite{wang2025NCFM}, to achieve enhanced performance on large-scale datasets. 

\noindent\textbf{Dataset pruning and condensation for ISR.} For ISR, multiple contributions have targeted the generation of a small subset of representative samples, based on diversity criteria such as texture complexity and blockiness distributions \cite{ding2023not, ohtani2024rethinking}. While applying data distillation/condensation could further enhance training efficiency, it is difficult to directly apply the techniques mentioned above to the ISR task, given the distinct differences between image classification and ISR \cite{liu2021discovering}. To the best of our knowledge, there has been very limited investigation into this research topic, with only one notable work \cite{zhang2024gsdd}, which utilizes GAN-based pretrained models to generate synthetic training images. However, this is still only applicable to an SR dataset with labels \cite{OST}.


\section{Method}

\subsection{Preliminaries}  
\label{sec:Preliminaries}
        
\textbf{Data condensation for ISR.} The training of ISR models employs paired high- and low-resolution (HR and LR) image patches, typically with a spatial resolution larger than 192$\times$192 (for HR patches) \cite{lim2017enhanced, liang2021swinir,  guo2024mambairv2}. However, most current mainstream dataset condensation methods proposed for high-level vision tasks are designed and validated on much lower resolution content (e.g., 32$\times$32 or 64$\times$64) \cite{wang2018dataset, zhao2021dataset, wang2025NCFM}. As a result, there are significant challenges when these methods are directly applied to the ISR task. First, training with high-resolution images that contain more pixels requires significantly increased optimization space (e.g., features, gradients, etc.), resulting in a much slower (or even impossible) training process. This is in particular relevant to the dataset condensation methods based on performance or gradient matching. Secondly, existing DC methods often rely on class labels to calculate task losses (such as the cross-entropy loss, soft labels, etc.) to guide the optimization of synthetic samples, while commonly used datasets in ISR tasks (e.g., DIV2K \cite{agustsson2017ntire}) are not associated with real class labels.

In this paper, our approach is based on distribution matching, which minimizes the discrepancy of distributions between two datasets. Specifically, given a real, original dataset $\mathcal{T}$ to a synthetic dataset $\mathcal{S}$, $ |\mathcal{S}| \ll |\mathcal{T}|$. The optimization goal is:
%
\begin{equation}
    \mathcal{S} = \underset{\mathcal{S}}{\mathrm{argmin}} \ \mathcal{L}(\mathcal{S},\mathcal{T}),
\end{equation}
where $\mathcal{L}$ stands for the distance measurement function. 

\vspace{5pt} \noindent\textbf{Distance measurement functions.} Early approaches \cite{wang2022cafe, zhao2023dataset} employ Mean Squared Error or Maximum Mean Discrepancy as $\mathcal{L}$, while recently, Neural Characteristic Function Discrepancy (NCFD) \cite{wang2025NCFM} has been proposed to capture distributional discrepancies by aligning the phases and amplitudes of neural features in the complex plane, achieving a balance between realism and diversity in the synthetic samples. Specifically, the optimization process is described by:

\vspace{-5pt}

\small
\begin{equation}
\underset{D_\mathcal{S}}{\min} \, \underset{\psi}{\max} \ \mathcal{L}_{dist}(D_\mathcal{T}, D_\mathcal{S}, f, \psi) = \underset{D_\mathcal{S}}{\min} \, \underset{\psi}{\max} \ \mathbb{E}_{\boldsymbol{x} \sim D_\mathcal{T}, \hat{\boldsymbol{x}} \sim D_\mathcal{S}} \int_{t} \sqrt{\mathrm{Chf}(\boldsymbol{t}; f)} \, dF(\boldsymbol{t}; \psi),
\end{equation}

\vspace{-5pt}

\begin{equation}
\begin{array}{ll}
\mathrm{Chf}(\boldsymbol{t}; f) = & \alpha \underbrace{ \left( \left( \left| \Phi_{f(\boldsymbol{x})}(\boldsymbol{t}) - \Phi_{f(\hat{\boldsymbol{x}})}(\boldsymbol{t}) \right| \right)^2 \right)}_{\textit{amplitude difference}} \\ 
  &+ \underbrace{(1 - \alpha) \cdot 
(2 \left| \Phi_{f(\boldsymbol{x})}(\boldsymbol{t}) \right| \left| \Phi_{f(\hat{\boldsymbol{x}})}(\boldsymbol{t}) \right|) \cdot (1 - \cos(\boldsymbol{a}_{f(\boldsymbol{x})}(\boldsymbol{t}) - \boldsymbol{a}_{f(\hat{\boldsymbol{x}})}(\boldsymbol{t})))}_{\textit{phase difference}}.
\end{array}
\end{equation}
\normalsize

\begin{figure}[t]
            \centering
            \includegraphics[width=1\linewidth]{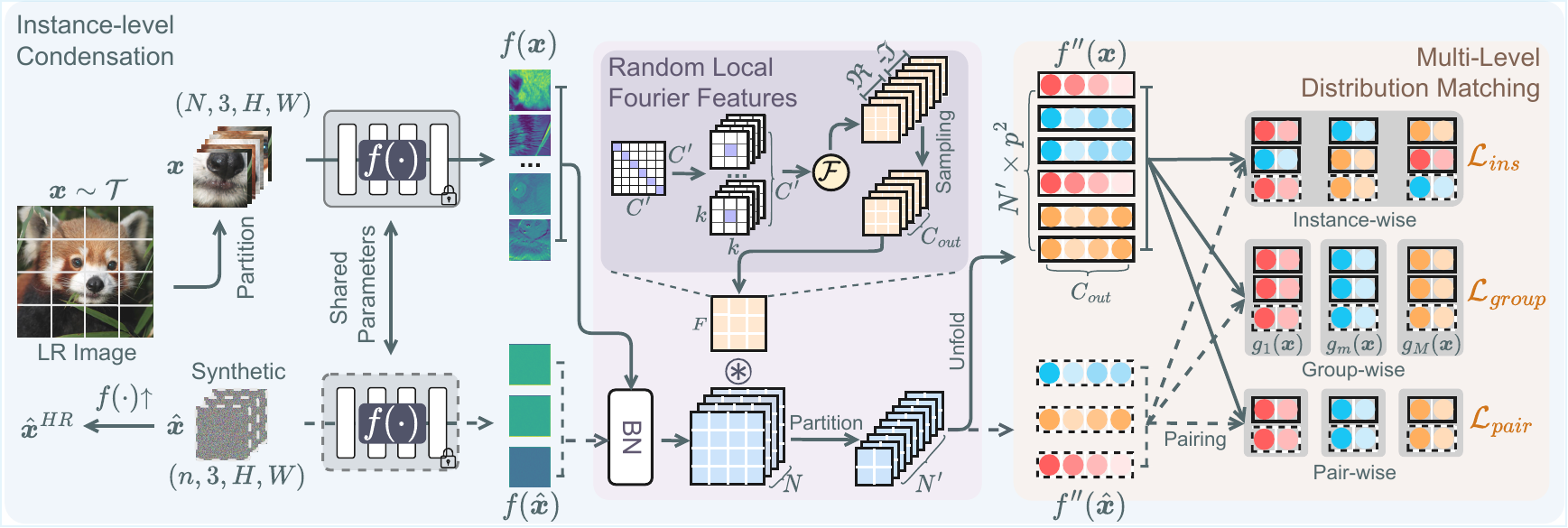}
            \caption{Illustration of the proposed Instance Data Condensation (IDC) framework.}
            \label{fig:framework}
            \vspace{-10pt}
\end{figure}   

\noindent Here $D$ denotes the distribution of a dataset. $f$ is the feature extractor that maps input training samples, i.e., $\boldsymbol{x}\in \mathcal{T}$ or $\hat{\boldsymbol{x}}\in\mathcal{S}$, into the latent space. $F(\boldsymbol{t},\psi)$ is the cumulative distribution function of the frequency argument $\boldsymbol{t}$, while $\psi$ is a parameterized sampling network used to obtain the distribution of $\boldsymbol{t}$. $\Phi_{f(\boldsymbol{x})}(\boldsymbol{t}) = \mathbb{E}_{f(\boldsymbol{x})} \left[ e^{j \langle \boldsymbol{t}, f(\boldsymbol{x}) \rangle} \right]$ stands for the characteristic function. $\mathrm{Chf}(\boldsymbol{t};f)$ calculates the distributional discrepancy between training and synthesized samples in the complex plane - here \textit{phase} $\boldsymbol{a}_{f(\boldsymbol{x})}(\boldsymbol{t})$ stands for data centers, which is crucial for realism, and \textit{amplitude} $\left| \Phi_{f(\boldsymbol{x})}(\boldsymbol{t}) \right|$ captures the distribution scale, contributing to the diversity. $\alpha$ is a hyper-parameter to balance the amplitude and phase information during the optimization process.

It is noted that while NCFD \cite{wang2025NCFM} has contributed to the SOTA dataset condensation method for image classification, it cannot capture useful features when directly applied to condense training datasets for ISR. This is illustrated in  \autoref{fig:IDC_feature}. This may be because learning the high-dimensional feature distribution is intractable, if we directly model the joint distribution of features with dimensions $C\times H\times W$. Furthermore, it uses a random Gaussian matrix for mapping features - $ f(\cdot)^{N \times C\times H\times W} \mapsto f(\cdot)^{N \times D}$ (when the sampling network is not used \cite{wang2025NCFM}) - (i) the random Gaussian projection fuses information globally, which, however, is unsuitable for many low-level vision tasks including ISR, where fine-grained local information is crucial; (ii) it does not capture the high-frequency features within high-resolution feature maps extracted by ISR models, thus limiting the capacity to learn fine details in the synthesis process.

\subsection{Instance Data Condensation} 
To address the unique challenges of the ISR task, we propose a novel \textbf{Instance Data Condensation (IDC)} framework. A critical bottleneck in applying standard dataset condensation to low-level vision is the heavy reliance on class labels, which are inherently absent in typical ISR datasets. IDC overcomes this issue by operating at the \textit{instance level}—treating each image as a ``class'' — thus making it suitable for \textbf{unlabeled datasets}. Furthermore, to efficiently handle high-resolution patches and preserve high-frequency details, IDC introduces a hierarchical \textbf{multi-level feature distribution loss}. This design performs local distribution matching across three progressive stages: (1) \textit{instance-level} (${\mathcal{L}}_{ins}$), which aligns the overall feature distribution of a single image to capture coarse visual structures; (2) \textit{group-level} (${\mathcal{L}}_{group}$), which partitions and clusters feature patches by similarity to learn fine-grained visual semantics; and (3) \textit{pair-wise level} (${\mathcal{L}}_{pair}$), which directly minimizes the discrepancy between each synthesized patch and its most similar real counterpart to ensure detail fidelity.

However, directly applying existing distribution matching techniques, such as the random Gaussian projection used in NCFM, is fundamentally incompatible with our proposed multi-level framework. Specifically, using this standard projection has two critical limitations for ISR: first, it performs a global transformation that destroys spatial structures, making the extraction and matching of local features impossible; secondly, it struggles to capture the high-frequency components that are essential for modeling fine textures in HR images. To solve these problems, we propose a novel \textbf{Random Local Fourier Features (RLFF)}. By transforming features into the spatial-frequency domain, RLFF explicitly captures rich high-frequency details while preserving the spatial layout. This dual capability effectively bridges the gap, enabling our multi-level distribution matching to be performed locally and ensuring high-fidelity detail synthesis.

As illustrated in \autoref{fig:framework}, the proposed IDC framework generates synthetic training pairs via a two-stage process. In the \textbf{first stage}, we synthesize the low-resolution (LR) training patches. Following our instance-level paradigm, given a real dataset $\mathcal{T}$ containing $I$ high-resolution (HR) images and their LR counterparts, we process each image individually. For a real LR image, its $N$ cropped patches, denoted as $\boldsymbol{x} \in \mathbb{R}^{N \times 3 \times H \times W}$, serve as the target distribution. We initialize a set of learnable synthetic patches $\hat{\boldsymbol{x}} \in \mathbb{R}^{n \times 3 \times H \times W}$, where $n = N \times r$ is defined by the condensation ratio $r \in (0,1)$. Using the feature extractor $f$ pretrained on the real dataset, we map both sets into the latent space. We then apply our proposed RLFF transformation to extract fine-grained local features, yielding $f'(\boldsymbol{x})$ and $f'(\hat{\boldsymbol{x}})$. Finally, the synthetic patches $\hat{\boldsymbol{x}}$ are optimized by minimizing the multi-level feature distribution loss mentioned above. 

In the \textbf{second stage}, we generate the corresponding HR targets. Since the matching framework exclusively synthesizes data in the LR input space, these patches inherently lack true HR counterparts. To resolve this, we employ the teacher ISR model pre-trained on the full dataset to up-sample the optimized $\hat{\boldsymbol{x}}$ into HR patches $\hat{\boldsymbol{x}}^{HR}$. This step acts as knowledge distillation \cite{hinton2015distilling}, providing regularized targets that guide the student model to learn robust features.

\vspace{-5pt}

{
\begin{algorithm}[t]
\captionsetup{font=small}
\caption{Instance Data Condensation. \label{alg:algorithm}}

\small 
\linespread{0.95}\selectfont 
\setlength{\interspacetitleruled}{0pt} 
\setlength{\interspacealgoruled}{0pt}  
\SetKwInput{KwInput}{Input} 
\SetKwInput{KwOutput}{Output}

\KwInput{Training dataset $\mathcal{T}=\{[\boldsymbol{x}_i, \boldsymbol{x}^{HR}_i ], i=1,\dots, I\}$; Synthetic dataset $\mathcal{S} = \varnothing$; ISR model $f$ pre-trained on $\mathcal{T}$; instance number $I$; condense ratio $r$; instance-level distribution loss weight $w_{ins}$; group-level distribution loss weight $w_{group}$; pair-wise loss weight $w_{pair}$; learning rate $\eta$. }

\For{ each real patch $\boldsymbol{x}_i$ in $\mathcal{T}$}{
    Randomly initialize the synthetic patches $\hat{\boldsymbol{x}}_i$, subject to $|\hat{\boldsymbol{x}}_i| = r |\boldsymbol{x}_i| $\;
     Extract the feature maps using $f$ for $\boldsymbol{x}_i$\;
    \For{$j$ in num\_iters}{
         Extract the feature maps using $f$ for $\hat{\boldsymbol{x}}_i$\;
        Extract their Random Local Fourier features and \textit{Unfold} the feature maps into local patches\;  
        \If{$j$ in warm up}{
            Compute $\mathcal{L} = w_{ins}\mathcal{L}_{ins}$\;
        }
        \Else{
        \If{$j$ in assigning}{
            Increase \textit{grouping} and \textit{pairing}\;
            }
            Compute $\mathcal{L} = w_{ins}  \mathcal{L}_{ins} +  w_{group} \mathcal{L}_{group} +  w_{pair} \mathcal{L}_{pair}$ \;
        }  \myrightcommentt{\autoref{sec:3.4}} 
        
        Update $\hat{\boldsymbol{x}}_i \leftarrow \hat{\boldsymbol{x}}_i - \eta\nabla_{\hat{\boldsymbol{x}}_i}\mathcal{L} $\;
    } \myrightcomment{\autoref{sec:3.3}}
    
    $\hat{\boldsymbol{x}}_i^{HR} = f(\hat{\boldsymbol{x}}_i)\uparrow$\;
    
    $\mathcal{S} = \mathcal{S} \cup \{[
\hat{\boldsymbol{x}}_i,\hat{\boldsymbol{x}}_i^{HR}]\} $\;
}
\KwOutput{$\mathcal{S}$}

\end{algorithm}
}
\noindent 
\subsection{Random Local Fourier Features}
\label{sec:3.3}

Given that the extracted feature maps of the real training patches are denoted as $f(\boldsymbol{x}) \in \mathbb{R}^{N \times c \times h \times w}$, in which $c,h,w$ are the channel number, size of the feature map, respectively, we first define an identity matrix $\mathds{1}_{C' \times C'}$ and reshape it into a convolutional filter, $E \in \mathbb{R}^{C' \times c \times k \times k}$ where $C' = c \times k^2$, and $k$ is the kernel size. This filter is used to map  $f(\boldsymbol{x})$ from the channel-spatial domain to the channel domain to extract local features, while keeping the spatial structure information. To extract the high frequency details, we further apply the Fourier transform $\mathcal{F}$ after extracting local features by $E$, which, in practice, is achieved by directly applying $\mathcal{F}$ to the convolutional filter $E$ in the output channel dimension, and decomposing the real and imaginary parts to form a new Fourier-based convolution filter $F\in \mathbb{R}^{2C' \times c \times k \times k}$:
\begin{equation}
F = [\Re(\mathcal{F}(E) ), -\Im(\mathcal{F}(E))].
\end{equation}
The feature maps, $f(\boldsymbol{x})$, are transformed by $F$ into a frequency-aware representation, $f'(\boldsymbol{x})$, which fully encodes the local spatial and channel information in the spatial-frequency domain that is particularly suitable for data and textures with periodic structures.

\begin{equation}
    f'(\boldsymbol{x})= f(\boldsymbol{x}) \circledast F.
\end{equation}

Here, to reduce the size of $f'(\boldsymbol{x})$ and the complexity for loss computation, we implement channel-wise random sampling to obtain $F \in \mathbb{R}^{C_{out} \times c \times k \times k}, C_{out}<C'$ \cite{wang2025NCFM}. We also apply batch normalization to $f(\boldsymbol{x})$ before convolution.

Finally, we partition $f'(\boldsymbol{x})$ to obtain local feature patches
 $ f''(\boldsymbol{x}) \in \mathbb{R}^{N' \times C_{out} \times p \times p }$, where $N'= N * \lceil h/p \rceil * \lceil w/p \rceil$, and $p$ is the patch size. In order to learn local features and make the high-dimensional distribution matching tractable, we further \textit{unfold} $ f''(\boldsymbol{x})$ from $\mathbb{R}^{N' \times C_{out} \times p \times p} $ to $\mathbb{R}^{(N' \times p \times p) \times C_{out}}$, which treats features in the patches as different samples in a batch. This decomposes the intractable global distribution modeling into manageable local patch distributions.

This operation (RLFF) has also been applied to the feature maps of the synthetic patches, $f(\hat{\boldsymbol{x}})$, before performing the multi-level distribution matching detailed below. The random sampling utilizes the same seed for both $f({\boldsymbol{x}})$ and $f(\hat{\boldsymbol{x}})$ at each training step, ensuring the frequency components are matched.

 \vspace{-5pt}
\subsection{Multi-level distribution matching}
\label{sec:3.4}
To adapt to the nature of the ISR task, rather than directly using existing distribution matching losses, we propose a Multi-level Distribution Matching approach to optimize the feature distribution of synthetic patches at both instance (class) and group levels and match pair-wise feature patches. 

\vspace{5pt}
\noindent\textbf{Matching instance-level feature distributions.} After obtaining local features for both real and synthetic patches, $f''(\boldsymbol{x})$ and $f''(\hat{\boldsymbol{x}})$, the instance-level feature distribution loss, $\mathcal{L}_{ins}$, is calculated to minimize the distributional discrepancy between the real and synthetic patches. Here we do not use the sampling network mentioned in its original literature \cite{wang2025NCFM} for efficient computation:
\vspace{-5pt}
\begin{equation}
\mathcal{L}_{ins}=\mathcal{L}_{dist}(\boldsymbol{x},\hat{\boldsymbol{x}}, f) 
= \mathbb{E}_{\boldsymbol{x}, \hat{\boldsymbol{x}}} \int_{\boldsymbol{t}} \sqrt{\text{Chf}(\boldsymbol{t}; f)} \, dF(\boldsymbol{t} ).
\end{equation}

\noindent\textbf{Matching group-wise feature distributions.}
While instance-level feature distribution matching ensures consistency in the overall (global) feature distribution between synthetic and original patches, local features often exhibit significant complexity and diversity, preventing instance-level matching from sufficiently capturing all distributional differences. To address this issue, we designed a more fine-grained, group-level feature distribution matching loss,  $\mathcal{L}_{group}$, which first partitions the real local features $f''(\boldsymbol{x})$ into $M$ groups using K-means clustering. Each synthetic local feature is then iteratively assigned to its nearest group centroid with a progressive assigning strategy. Here, the number of synthetic features assigned to each group is proportional to the number of real features in that group, and the assignment is updated in steps to maintain stable optimization. To ensure that the assignments of neighboring features are consistent and to reduce complexity, we perform grouping in the patch level, i.e., features in the same patch are averaged and assigned into the same group. The resulting set of grouped local features is denoted as $g_m(\boldsymbol{x})$ and $g_m(\hat{\boldsymbol{x}}), m=1,\dots,M$.

After this assignment, we compute the feature distribution matching loss for each group to more accurately learn the synthetic data, by reflecting the real data distribution at the group level. This is described by the following equation:
\vspace{-5pt}
\begin{equation}
    \mathcal{L}_{group} = \sum_{m=1}^M \mathcal{L}_{dist}(g_m(\boldsymbol{x}),g_m(\hat{\boldsymbol{x}}),id(\cdot)),
\end{equation}
in which $id$ denotes the identity function, i.e. $id(x)=x$.

\vspace{5pt}
\noindent\textbf{Matching pair-wise features.}
Moreover, to further address the unique nature of the ISR task, which aims to reconstruct content with high fidelity and fine local details, we also introduce a pair-wise loss, $\mathcal{L}_{pair}$, within each local feature group described above. Specifically, for each synthetic feature patch assigned to a group, we identify the most similar real local feature patch within the same group and construct a pair. We then compute the L1 loss between the paired features to minimize their discrepancy. This is expected to encourage the synthetic data to better match the real data at the local detail level and to improve the fidelity of fine details and the overall visual quality of the synthesized images. This process can be described by the equation below:
\begin{equation}
    \mathcal{L}_{pair} = \sum_{m=1}^M \sum_{i=1}^{N_m} \frac{1}{N_m}  \| f''(\boldsymbol{x})^{g_m(i)} - f''(\hat{\boldsymbol{x}})^{g_m(i)} \|_1,
\end{equation}
in which $N_m$ is the number of feature patch pairs in the group $m$. $f''(\boldsymbol{x})^{g_m(i)}$ represents the real feature in the $i$\textsuperscript{th} pair, and $f''(\hat{\boldsymbol{x}})^{g_m(i)}$ is the corresponding synthetic feature in this pair.


These three losses are used at various stages in the proposed method for synthetic data optimization, and the training algorithm is described in \autoref{alg:algorithm}.

\section{Experiment Configuration}

\textbf{Implementation details.} IDC is a novel dataset condensation method specifically designed for the ISR task, which can be applied to any dataset with or without labels. In this experiment, we employ SwinIR \cite{liang2021swinir} and MambaIRv2 \cite{guo2024mambairv2}, which are optimized on the original real training dataset $\mathcal{T}$ obtained from DIV2K as the feature extractor and the up-sampling ISR model, respectively. For each instance/class, we can generate the synthetic dataset with a 10\% condense ratio for 20k iterations with a single Nvidia mid-range performance GPU. For HR patch up-sampling, we choose one of the latest ISR models, MambaIRv2. Here, the outputs of the up-sampling ISR models (teacher) act as a form of knowledge distillation, providing regularized targets that guide the model to learn more generalizable features. Other training and hyper-parameter configurations based on the ablation study are also provided in \textit{Sec 3.1 of Supplementary}.

\vspace{5pt} \noindent\textbf{Datasets and metrics.} To evaluate the performance of the proposed method, we used the widely used training dataset for ISR, DIV2K \cite{agustsson2017ntire}, as the original training dataset; this contains 800 images with a 2K resolution. We follow the common practice \cite{basicsr,lim2017enhanced, zhang2018image,wang2022uformer, liang2021swinir, shi2025vmambair} to crop all the images into overlapped patches with a spatial resolution of 256$\times$256 and down-sample them into corresponding LR patches (128$\times$128 and 64$\times$64), targeting the $\times$2 and $\times$4 tasks. This results in a total of 120,765 HR-LR patch pairs, forming the real dataset $\mathcal{T}$ mentioned above. We use five commonly used datasets \cite{Set5, Set14, Urban100, martin2001database, fujimoto2016manga109} for performance evaluation and measured by PSNR and SSIM \cite{wang2004image} on RGB channels.

\vspace{5pt}
\noindent\textbf{ISR models.} Three popular ISR models, including EDSR-baseline \cite{lim2017enhanced}, SwinIR \cite{liang2021swinir}, and MambaIRv2 \cite{guo2024mambairv2}, have been trained here on different datasets generated by the proposed and benchmark methods in this experiment, based on the same training configurations. To ensure a fair comparison, we re-trained all models on the full dataset and other baselines from scratch. The training and evaluation configurations for ISR methods are summarized in \textit{Sec 2.3 of Supplementary}.

\vspace{5pt}\noindent\textbf{Baseline Methods.} We compared the proposed methods with four dataset core-selection/pruning methods, including random selection, Herding \cite{welling2009herding}, Kcenter \cite{Kcentergreedy} and DCSR \cite{ding2023not}. All these benchmarks were applied on $\mathcal{T}$ with the same ratio (10\%) to obtain the selected/pruned (real) patches and the implementation details are provided in \textit{Sec 2.2 of Supplementary}. We also provide the results based on the original training set $\mathcal{T}$ (Whole) for reference. We did not compare IDC to the only available dataset condensation method designed for ISR, GSDD \cite{zhang2024gsdd}, because it is only suitable for datasets with classification labels and was evaluated with GAN-based ISR models, which is not a common practice in ISR literature. Moreover, we did not directly benchmark our approach against DC methods for high-level vision tasks due to their label-based nature. However, we do take them into account in the ablation study below. 

\vspace{-5pt}

\begin{table}[!t]
\centering
\footnotesize
\caption{Comparison with coreset selection and dataset pruning methods. For all methods (except the Whole), we generated/selected 12,076 LR-HR pairs with the condense ratio $r$=10\%. The results are based on PSNR (dB) (SSIM results are shown in \textit{Sec 4.1 of Supplementary}). The best and second best results are highlighted in  \textcolor{red}{red} and \textcolor{blue}{blue}, respectively.}
\vspace{5pt}
\resizebox{\linewidth}{!}{%
\begin{tabular}{r r R{1cm}R{1cm}R{1cm}R{1cm}R{1cm}R{1cm}R{1cm}R{1cm}R{1cm}R{1cm}R{1cm}}
\toprule
& \multirow{2}{*}{PSNR (dB)$\uparrow$} 
& \multicolumn{2}{c}{Set5 \cite{Set5}} & \multicolumn{2}{c}{Set14 \cite{Set14}} & \multicolumn{2}{c}{Urban100 \cite{Urban100}} & \multicolumn{2}{c}{BSD100 \cite{martin2001database}} & \multicolumn{2}{c}{Manga109 \cite{fujimoto2016manga109}}\\
\cmidrule{3-12}
& & \multicolumn{1}{c}{$\times$2} & \multicolumn{1}{c}{$\times$4} & \multicolumn{1}{c}{$\times$2} & \multicolumn{1}{c}{$\times$4} & \multicolumn{1}{c}{$\times$2} & \multicolumn{1}{c}{$\times$4} & \multicolumn{1}{c}{$\times$2} & \multicolumn{1}{c}{$\times$4} & \multicolumn{1}{c}{$\times$2} & \multicolumn{1}{c}{$\times$4}\\
 \midrule
 \multirow{5}{*}{\rotatebox{270}{EDSR \cite{lim2017enhanced}}} & Whole  & \textcolor{red}{35.79} & \textcolor{blue}{30.17} & \textcolor{blue}{31.48} & \textcolor{blue}{26.61} & \textcolor{red}{30.41} & \textcolor{red}{24.50} & \textcolor{red}{30.83} & \textcolor{red}{26.24} & \textcolor{red}{36.13} & \textcolor{blue}{28.50}\\
 & Random  & 35.70 & 30.13 & 31.38 & 26.54 & 30.24 & 24.40 & 30.77 & 26.19 & 35.94 & 28.32\\
 & Herding \cite{welling2009herding} & 35.70 & 29.94 & 31.38 & 26.43 & 30.22 & 24.10 & 30.78 & 26.11 & 35.90 & 27.88\\
 & Kcenter \cite{Kcentergreedy} & 35.67 & 30.01 & 31.40 & 26.50 & 30.28 & 24.28 & 30.80 & 26.17 & 35.98 & 28.19\\
 & DCSR \cite{ding2023not} & 35.71 & 30.16 & 31.41 & 26.54 & 30.20 & 24.43 & 30.79 & \textcolor{blue}{26.21} & 35.99 & 28.42\\
 & \cellcolor{MyLightGreen}\textbf{IDC (ours)} & \cellcolor{MyLightGreen}\textcolor{blue}{35.73} & \cellcolor{MyLightGreen}\textcolor{red}{30.18} & \cellcolor{MyLightGreen}\textcolor{red}{31.55} & \cellcolor{MyLightGreen}\textcolor{red}{26.63} & \cellcolor{MyLightGreen}\textcolor{blue}{30.38} & \cellcolor{MyLightGreen}\textcolor{blue}{24.47} & \cellcolor{MyLightGreen}\textcolor{blue}{30.81} & \cellcolor{MyLightGreen}\textcolor{red}{26.24} & \cellcolor{MyLightGreen}\textcolor{blue}{36.08} & \cellcolor{MyLightGreen}\textcolor{red}{28.54}\\
 \midrule
  \multirow{5}{*}{\rotatebox{270}{SwinIR \cite{liang2021swinir}}} & Whole  & \textcolor{blue}{35.81} & 30.28 & \textcolor{blue}{31.56} & \textcolor{red}{26.78} & \textcolor{red}{30.63} & \textcolor{red}{24.92} & \textcolor{red}{30.86} & \textcolor{red}{26.34} & \textcolor{blue}{36.36} & \textcolor{blue}{28.98}\\
 & Random  & 35.79 & 30.20 & 31.53 & 26.68 & 30.49 & 24.70 & 30.84 & 26.28 & 36.27 & 28.74\\
 & Herding \cite{welling2009herding} & 35.77 & 30.03 & 31.52 & 26.54 & 30.51 & 24.40 & 30.84 & 26.19 & 36.23 & 28.30\\
 & Kcenter \cite{Kcentergreedy} & 35.74 & 30.12 & 31.54 & 26.61 & 30.55 & 24.56 & 30.85 & 26.25 & 36.28 & 28.64\\
 & DCSR \cite{ding2023not} & \textcolor{blue}{35.81} & \textcolor{blue}{30.33} & \textcolor{blue}{31.56} & \textcolor{blue}{26.69} & 30.60 & 24.78 & \textcolor{red}{30.86} & \textcolor{blue}{26.31} & 36.37 & 28.86\\
 &  \cellcolor{MyLightGreen}\textbf{IDC (ours)} & \cellcolor{MyLightGreen} \textcolor{red}{35.90} &  \cellcolor{MyLightGreen}\textcolor{red}{30.34} &  \cellcolor{MyLightGreen}\textcolor{red}{31.77} &  \cellcolor{MyLightGreen}\textcolor{red}{26.78} &  \cellcolor{MyLightGreen}\textcolor{blue}{30.86} &  \cellcolor{MyLightGreen}\textcolor{blue}{24.86} &  \cellcolor{MyLightGreen}\textcolor{red}{30.93} &  \cellcolor{MyLightGreen}\textcolor{red}{26.34} &  \cellcolor{MyLightGreen}\textcolor{red}{36.64} &  \cellcolor{MyLightGreen}\textcolor{red}{29.02}\\
 \midrule
  \multirow{6}{*}[3ex]{\rotatebox{270}{Mambairv2 \cite{guo2024mambairv2}}}& Whole  & \textcolor{red}{38.13} & \textcolor{red}{30.52} & \textcolor{blue}{33.80} & \textcolor{red}{26.88} & \textcolor{red}{32.85} & \textcolor{red}{25.21} & \textcolor{red}{32.31} & \textcolor{red}{26.40} & \textcolor{blue}{39.02} & \textcolor{blue}{29.22}\\
 & Random  & 38.10 & 30.49 & 33.64 & 26.86 & 32.50 & 25.02 & 32.26 & 26.37 & 38.94 & 29.07\\
 & Herding \cite{welling2009herding} & 38.11 & 30.36 & 33.60 & 26.65 & 32.47 & 24.65 & 32.25 & 26.30 & 38.82 & 28.54\\
 & Kcenter \cite{Kcentergreedy} & 38.06 & 30.45 & 33.68 & 26.79 & 32.56 & 24.86 & 32.27 & 26.34 & 38.92 & 28.89\\
 & DCSR \cite{ding2023not} & \textcolor{blue}{38.12} & \textcolor{blue}{30.51} & 33.74 & 26.86 & 32.63 & 25.10 & \textcolor{blue}{32.28} & \textcolor{blue}{26.39} & 38.98 & 29.14\\
 & \cellcolor{MyLightGreen}\textbf{IDC (ours)} & \cellcolor{MyLightGreen}\textcolor{red}{38.13} & \cellcolor{MyLightGreen}\textcolor{red}{30.52} & \cellcolor{MyLightGreen}\textcolor{red}{33.80} & \cellcolor{MyLightGreen}\textcolor{red}{26.91} & \cellcolor{MyLightGreen}\textcolor{blue}{32.80} & \cellcolor{MyLightGreen}\textcolor{blue}{25.16} & \cellcolor{MyLightGreen}\textcolor{red}{32.31} & \cellcolor{MyLightGreen}\textcolor{red}{26.40} & \cellcolor{MyLightGreen}\textcolor{red}{39.05} & \cellcolor{MyLightGreen}\textcolor{red}{29.26}\\
\bottomrule
\end{tabular}}
\label{tab:main_results}
\vspace{-10pt}
\end{table}

\section{Results and Discussion}

\begin{figure}[!t]
    \centering
    \begin{minipage}[b]{0.48\linewidth}
        \centering
        \includegraphics[width=\linewidth]{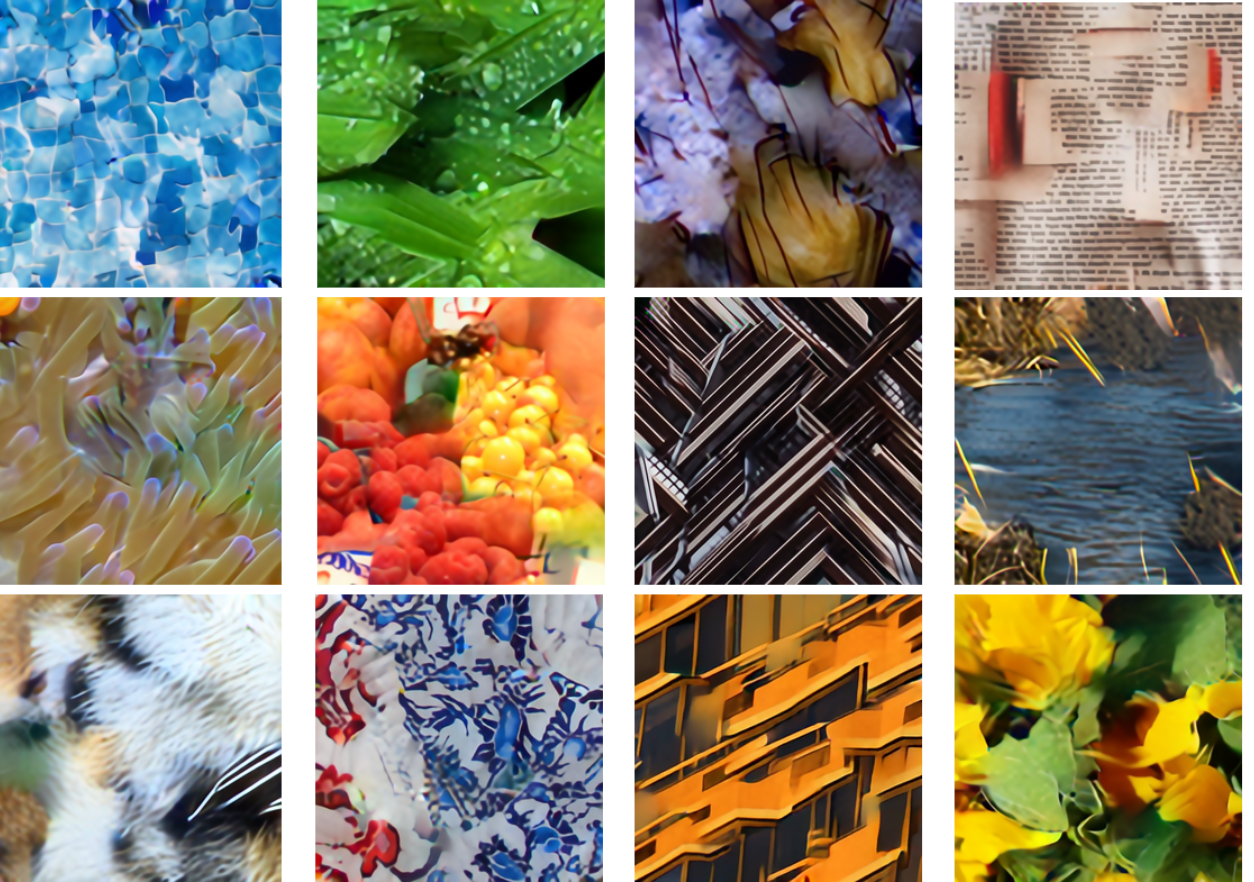}
    \end{minipage}
    \hfill
    \begin{minipage}[b]{0.50\textwidth}
        \centering
        \includegraphics[width=\linewidth]{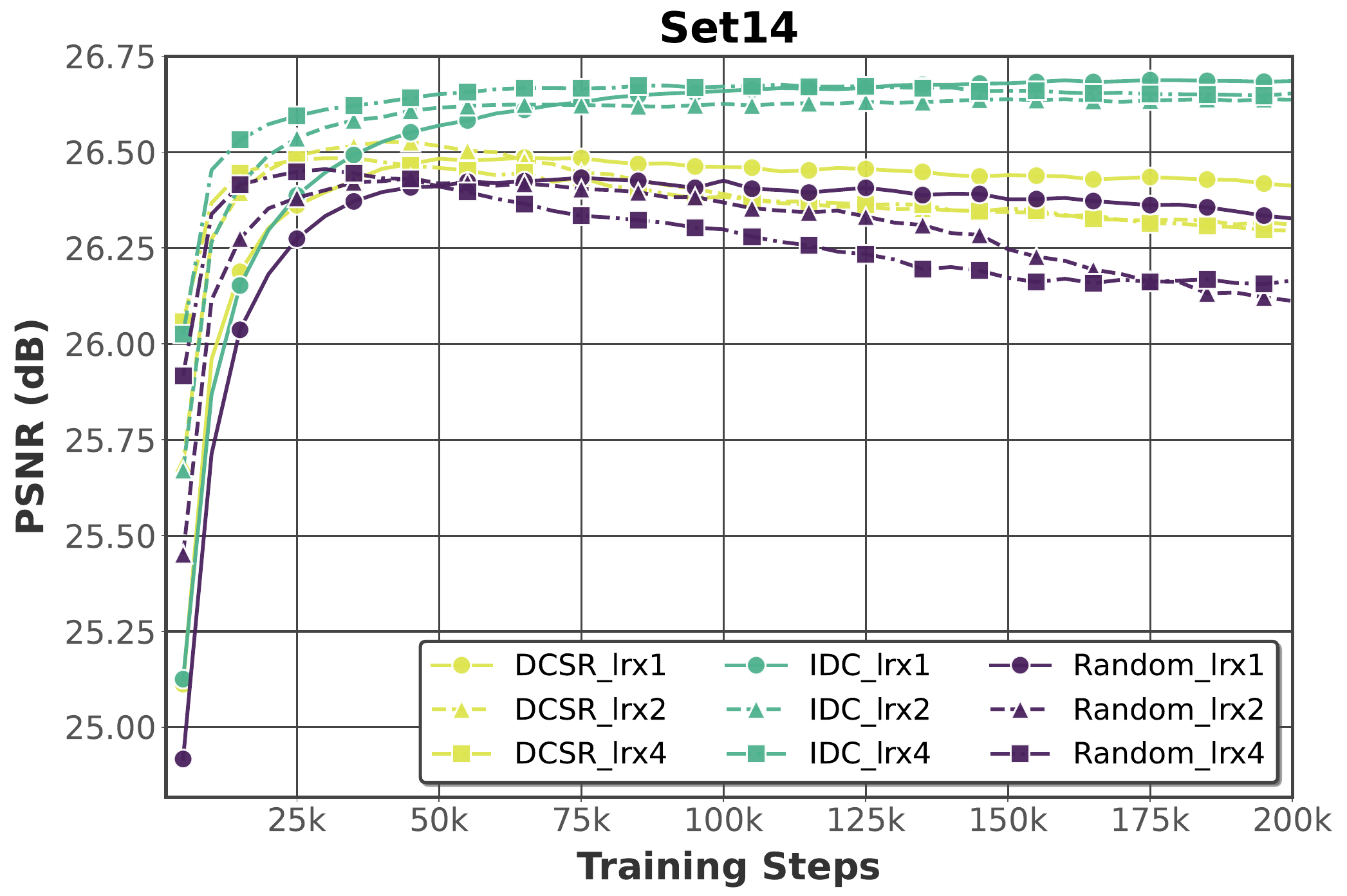}
    \end{minipage}
    \caption{\textbf{(Left)}: Visual examples of IDC synthesized images. \textbf{(Right)}: Validation trajectory on the Set14.}
    \label{fig:comparison}
    \vspace{-15pt}
\end{figure}

\textbf{Overall performance.} \autoref{tab:main_results} summarizes the quantitative results of our proposed IDC approach and other dataset selection/pruning methods for ISR, while the comparison of reconstruction quality for different ISR methods is provided in \textit{Sec 6.3 of Supplementary}. It can be observed that IDC consistently achieves superior performance compared to all the benchmark methods across all test datasets and quality metrics. In particular, with only 10\% of the data volume, it offers even better evaluation performance compared to the whole original training set on four out of five datasets. Moreover, we provide visual examples of synthetic training patches in \autoref{fig:comparison}, which show that IDC can preserve high-frequency details and texture information. As far as we are aware, this is the first data condensation method that achieves this level of performance for the ISR task.

\noindent\textbf{Scalability and Generalization.}
We further assess the scalability and generalization of our method by investigating two key considerations for practical dataset condensation: supporting a higher condensation ratio and preventing models from overfitting on small synthetic datasets. To this end, we adopted a more challenging 1\% condensation ratio (results in \textit{Sec 4.2 of Supplementary}), and tracked the test performance of the corresponding ISR models trained on different condensed datasets with different learning rates. As illustrated in \autoref{fig:comparison} (Right), even at this aggressive ratio, our method maintains a stable, upward learning curve throughout the training, whereas baseline methods exhibit clear signs of overfitting. We also plotted the training loss versus validation performance in \textit{Fig. 6 and 7 in Supplementary}, confirming stronger scalability and generalization of our framework.



\vspace{5pt}\noindent\textbf{Training efficiency and Cost-Benefit analysis.} IDC significantly accelerates the training pipeline. As shown in \autoref{tab:convergence_booktabs} and \textit{Fig. 8 of Supplementary}, models trained on our 10\% condensed dataset reach target PSNRs with 2-4 $\times$ fewer iterations than those trained on the full dataset, enabling efficient hyperparameter tuning. Although the initial condensation takes roughly 1.5 GPU-hours per instance, this cost is highly competitive with high-level vision methods like DC \cite{zhao2021dataset} (1.6 GPU-hours/class on CIFAR-10) and TESLA \cite{cui2023scaling} (10.5 GPU-hours/class on CIFAR-100), especially given our higher-resolution inputs and complex feature extractors. More importantly, condensation is a one-time, upfront investment; once generated, the synthetic dataset becomes a reusable asset that accelerates all subsequent training sessions. Combined with a 90\% storage reduction, these long-term efficiency gains strongly justify the one-time computational investment.

\begin{table}[!t]
\centering
    \centering
    \caption{Comparison of training iterations (in thousands, `k') required to reach \textbf{two} different target PSNR values for each benchmark on SwinIR model at $\times 4$ scale.}
    \label{tab:convergence_booktabs}
\vspace{5pt}
\resizebox{\linewidth}{!}{%

\begin{tabular}{r *{10}{R{1cm}}}
    \toprule
    \textbf{}  & \multicolumn{2}{r}{Set5 \cite{Set5}} & \multicolumn{2}{r}{Set14 \cite{Set14}} & \multicolumn{2}{r}{Urban100 \cite{Urban100}} & \multicolumn{2}{r}{BSD100 \cite{martin2001database}} & \multicolumn{2}{r}{Manga109 \cite{fujimoto2016manga109}} \\
    \midrule
    Whole & 20k & 60k & 15k & 50k & 45k & 135k & 10k & 35k & 45k & 125k \\
    Uniform & 20k & 60k & 15k & 55k & 45k & 180k & 10k & 35k & 50k & 150k \\
    IDC(ours) & \textbf{10k} & \textbf{20k} & \textbf{10k} & \textbf{20k} & \textbf{20k} & \textbf{55k} & \textbf{5k} & \textbf{15k} & \textbf{15k} & \textbf{40k} \\
    \midrule
    Target PSNR & 29.38 & 29.99 & 26.00 & 26.54 & 24.18 & 24.67 & 25.55 & 26.07 & 28.11 & 28.69\\
    \bottomrule
\end{tabular}
}
\end{table}

\vspace{5pt}\noindent\textbf{Ablation study.} To validate the efficacy of our proposed components, we conducted a systematic ablation study \autoref{table: ablation}. Starting from the original NCFD baseline, progressively integrating our proposed modules yields consistent performance gains. We explicitly analyzed the roles of RLFF and the Unfolding operation; while removing either component individually (variants v5 and v6) results in a moderate performance drop due to their partially overlapping benefits in feature modeling. Furthermore, we compared RLFF (v5.1) against other feature extraction methods which fit the requirement of the multi-level feature distribution matching, i.e., preserving high frequency details and the spatial layout of the input. We consider alternatives including: applying DCT on the convolution filter (v5.2) and Wavelets (v5.3) in \autoref{table: ablation}, where RLFF consistently demonstrates superior performance in capturing the fine-grained textures required for ISR. We also provide their synthetic patch examples of v1-v4 in \autoref{fig:ablation} for visualization. More visual examples are provided in \textit{Sec 6.1 of Supplementary}.

\noindent\textbf{Selection of the Teacher Model.}
In our framework, a pre-trained teacher model is used to generate HR patches from the synthesized LR versions. This is because directly synthesizing HR patches does not align with our condensation process, where the pre-trained feature extractor takes LR (rather than HR) as input. A possible concern is whether the teacher imposes a hard performance ceiling on student models trained on the condensed data. To investigate this, we conducted a study where both the teacher model (for generating HR patches) and the student model (for evaluation) are set to EDSR, SwinIR, or MambaIRv2 in a cross-architecture configuration. The results in \textit{Sec 3.2 of Supplementary} show that no hard performance ceiling exists. For example, a stronger teacher does not always yield a superior student; in some cases, a student model can even outperform another student trained with a more powerful teacher. This suggests that the teacher's outputs act as a form of regularization and will not cause the training outcome to collapse into the teacher model’s generation pattern. Moreover, a key advantage of our framework is that the computationally expensive condensation phase is entirely decoupled from the teacher model. This allows efficient updates using future superior models via a fast inference. 

\begin{table}[!t]
\centering
\footnotesize
\caption{Results of the ablation study. \checkmark: included, \xmark: excluded. Here, we only condense 80 images/classes for each setting but with the same condense ratio 10\% in each class (effectively 1\% condense ratio for the whole dataset). We choose the EDSR-baseline \cite{lim2017enhanced} as the ISR model, evaluated on three test datasets at $\times 4$ scale.}
\vspace{5pt}
\label{table: ablation}
\resizebox{\textwidth}{!}{
\begin{tabular}{rcccccrrrrrr}
\toprule
 \multirow{2}{*}{Variant}& \multirow{2}{*}{Local Feature} & \multirow{2}{*}{Unfolding} & \multirow{2}{*}{Instance Loss} & \multirow{2}{*}{Group Loss} & \multirow{2}{*}{Pair Loss}  & \multicolumn{2}{c}{Set5 \cite{Set5}}& \multicolumn{2}{c}{Urban100 \cite{Urban100}}& \multicolumn{2}{c}{Manga109 \cite{fujimoto2016manga109}} \\
\cmidrule{7-12}
&&&&& & PSNR & SSIM & PSNR & SSIM & PSNR & SSIM \\
\midrule
\textbf{IDC}  & \checkmark  & \checkmark  & \checkmark  & \checkmark  & \checkmark    &30.02  & 0.8616&24.07 & 0.7474& 28.00& 0.8723\\
v1 & \xmark   & \xmark   & \checkmark  & \xmark   & \xmark    &  -21.71&-0.7606 &-16.60&-0.6674 & -20.44&-0.7499   \\
v2 & \checkmark  & \xmark   & \checkmark & \xmark   & \xmark   &--0.49&-0.0092 &-0.60&-0.0247 & -0.56&-0.0108  \\
v3 & \checkmark  & \checkmark  & \checkmark & \xmark   & \xmark  &  -0.16 &-0.0024 &  -0.12 &-0.0059  &-0.04&-0.0015 \\
v4 & \checkmark  & \checkmark  & \checkmark  & \checkmark  & \xmark    &  -0.05 &-0.0008 & -0.04 &-0.0020  & 0.06 &0.0007   \\
v5 & \xmark   &\checkmark  & \checkmark  & \checkmark  & \checkmark  & -0.19&-0.0023 & -0.04&-0.0017 & -0.06&-0.0005 \\
v6 & \checkmark  & \xmark   & \checkmark  & \checkmark  & \checkmark   &-0.29  & -0.0034  &  -0.10  &-0.0048  & -0.10  &-0.0017\\
v7 & \checkmark  & \checkmark  & \xmark   & \checkmark  & \checkmark  & -0.22& -0.0032 & -0.02&-0.0012&-0.09&-0.0011  \\
\midrule
v5.1 &    w/o RLFF &\checkmark&\checkmark&\checkmark&\checkmark& -0.32 & -0.0059 & -0.23 & -0.0110 & -0.17 & -0.0044 \\
v5.2 &    DCT &\checkmark&\checkmark&\checkmark&\checkmark& -0.10 & -0.0017 & -0.05 & -0.0036 & -0.02 & -0.0009 \\
v5.3 &    Wavelets &\checkmark&\checkmark&\checkmark&\checkmark& -0.58 & -0.0107 & -0.64 & -0.0228 & -0.52 & -0.0095 \\
\bottomrule
\end{tabular}
}
\vspace{-10pt}
\end{table}

\begin{figure}[t]
    \centering
    \includegraphics[width=\linewidth]{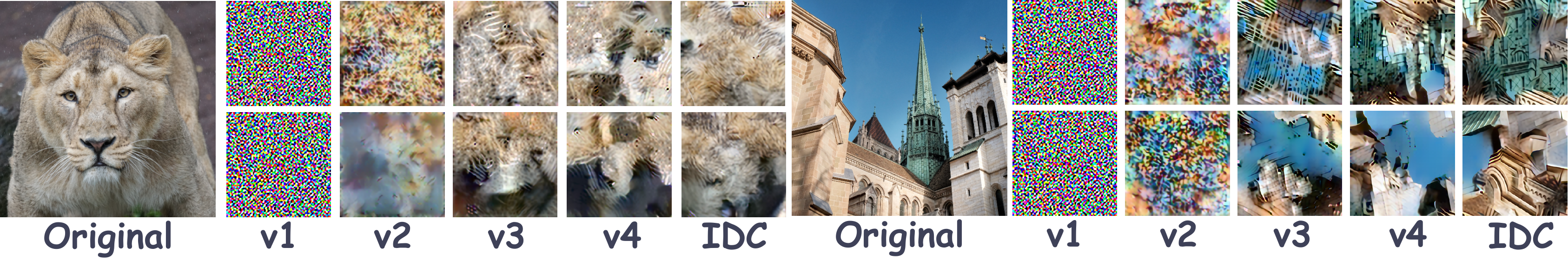}

    \caption{Visualization for each tested model variants.
    }
    \label{fig:ablation}
\vspace{-10pt}
\end{figure}

\begin{table}[!t]
\centering
\caption{Results on Flickr2K \cite{lim2017enhanced} dataset with condense ratio $r$=1\% and evaluated on SwinIR model at $\times 4$ scale. The results are based on PSNR (dB) and SSIM. }
\vspace{5pt}
\label{tab:flickr2k}
\resizebox{\linewidth}{!}{%
\begin{tabular}{r c c c c c c c c c c c}
\toprule
\multirow{2}{*}{} &Condense  & \multicolumn{2}{c}{Set5 \cite{Set5}} & \multicolumn{2}{c}{Set14 \cite{Set14}} & \multicolumn{2}{c}{Urban100 \cite{Urban100}} & \multicolumn{2}{c}{BSD100 \cite{martin2001database}} & \multicolumn{2}{c}{Manga109 \cite{fujimoto2016manga109}} \\
\cmidrule{3-12}
 & Ratio& PSNR & SSIM & PSNR & SSIM & PSNR & SSIM & PSNR & SSIM & PSNR & SSIM \\
\midrule
Whole(Flickr2K) & - & 30.39 & 0.8688 & 26.85 & 0.7505 & 24.97 & 0.7779 & 26.36 & 0.7193 & 29.13 & 0.8891 \\

Random & 1\% & 30.19 & 0.8659 & 26.57 & 0.7444 & 24.45 & 0.7634 & 26.22 & 0.7151 & 28.53 & 0.8823 \\

DCSR & 1\% & 30.24 & 0.8662 & 26.63 & 0.7453 & 24.49 & 0.7652 & 26.24 & 0.7160 & 28.60 & 0.8831 \\

\rowcolor{MyLightGreen} 
\textbf{IDC (ours)} & 1\% & \textbf{30.34} & \textbf{0.8678} & \textbf{26.75} & \textbf{0.7482} & \textbf{24.84} & \textbf{0.7738} & \textbf{26.35} & \textbf{0.7183} & \textbf{29.01} & \textbf{0.8878} \\
\bottomrule
\end{tabular}%
}
\vspace{-5pt}
\end{table}

\begin{table}[!t]
\centering
\caption{Results on color image denoising ($\sigma=50$). `Whole' represents a large-scale training dataset (8,594 images) and all methods are evaluated on the SwinIR model.}
\label{tab:denoising}
\vspace{5pt}
\setlength{\tabcolsep}{6pt} 
\small
\begin{tabular}{r c c c c c c c c c}
\toprule
\multirow{2}{*}{} &Condense  & \multicolumn{2}{c }{McMaster\cite{zhang2011color}} & \multicolumn{2}{c }{Kodak24\cite{Franzen1999Kodak}} & \multicolumn{2}{c}{CBSD68\cite{martin2001database}} & \multicolumn{2}{c}{Urban100\cite{Urban100}}   \\
\cmidrule{3-10}
 &Ratio& PSNR&SSIM & PSNR&SSIM & PSNR&SSIM& PSNR&SSIM \\
\midrule
Whole  &- & 35.53 & 0.9335 & 35.28 & 0.9292 & 34.37 & 0.9351  & 35.04 & 0.9516 \\
Uniform
&10\%&  35.48 & 0.9328 & 35.26 & 0.9288 & 34.35 & 0.9347  & 34.99 & 0.9510  \\
\textbf{IDC (ours)}  

&1\%& \textbf{35.51} & 0.9316 & \textbf{35.26} & 0.9291 & \textbf{34.36} & 0.9348 & \textbf{35.00} & 0.9506  \\

\bottomrule
\end{tabular}
\vspace{-15pt}
\end{table}

\noindent\textbf{Evaluation on the larger dataset.}
\label{sec:larger} To further validate the robustness of our framework, we conducted experiments on Flickr2K \cite{lim2017enhanced}, a larger scale training dataset for ISR (2650 images). As presented in \autoref{tab:flickr2k}, our IDC method, even at an aggressive 1\% condensation ratio, consistently outperforms both Random Selection and DCSR baselines across all metrics. More Remarkably, our 1\% condensed set remains competitive with the Whole dataset.

\noindent\textbf{Evaluation of the image denoising task.}
We validated the task and dataset extensibility by applying IDC to a large-scale color image denoising dataset of 8,594 images\footnote{This dataset collects DIV2K \cite{agustsson2017ntire}, Flickr2K\cite{lim2017enhanced}, BSD500\cite{amfm_pami2011}, and WED\cite{ma2016waterloo} by following the standard protocol \cite{liang2021swinir}.}. Results in \autoref{tab:denoising} show that our 1\% condensed set achieves performance comparable to a 10\% baseline subset - this confirms the extensibility of the proposed method to other low-level vision tasks.

\section{Conclusion}

This paper presents Instance Data Condensation (IDC) specifically targeting image super-resolution. It is designed to synthesize a small yet informative training dataset from a large dataset containing real images. By leveraging a multi-level distribution matching framework and the new Random Local Fourier Features, IDC captures essential structural and textural features from the original high-resolution images and achieves significant data condensation. Trained on the resulting small synthetic dataset (with only 10\% of the original data volume) ISR models can achieve comparable performance to the entire real training dataset (DIV2K), when benchmarked on multiple test datasets. As far as we are aware, this is the first ISR data condensation approach offering this level of performance.  More importantly, compared to other benchmarks based on data selection and pruning methods, it provides better training stability and faster training convergence. The proposed instance-level paradigm may also inspire new approaches for data condensation in other unlabeled, low-level vision tasks. We recommend that future work should focus on further performance improvement and speeding up the condensation process.

{
\small
\bibliographystyle{abbrv}
\bibliography{main}
}

\end{document}